\documentclass{article}

\PassOptionsToPackage{numbers}{natbib}
\usepackage[final]{nips_2016}

\usepackage[utf8]{inputenc} 
\usepackage[T1]{fontenc}    
\usepackage{hyperref}       
\usepackage{url}            
\usepackage{booktabs}       
\usepackage{amsfonts}       
\usepackage{nicefrac}       
\usepackage{microtype}      

\usepackage{todonotes}
\usepackage{mathtools}
\usepackage{amssymb}
\usepackage{amsmath}
\usepackage{bm}
\usepackage{algorithm}
\usepackage{algorithmicx} 
\usepackage{algpseudocode}
\usepackage{caption}
\usepackage{subcaption}
\usepackage{listings}
\usepackage{placeins}
\usepackage{framed}
\usepackage[]{ntheorem}
\usepackage{enumitem}

\usepackage{adjustbox}
\usepackage{etoolbox}

\makeatletter
\expandafter\patchcmd\csname\string\algorithmic\endcsname%
      {\labelwidth 0.5em}{\labelwidth0pt\labelsep0pt}{}{}

\algnewcommand{\LineComment}[1]{\State \(\triangleright\) #1}
\algnewcommand{\LineCommentx}[1]{\Statex \(\triangleright\) #1}
\algnewcommand{\IIf}[1]{\State\algorithmicif\ #1\ \algorithmicthen}
\algnewcommand{\EndIIf}{\unskip\ \algorithmicend\ \algorithmicif}
\algnewcommand{\FFor}[1]{\State\algorithmicfor\ #1\ \algorithmicdo}
\algnewcommand{\EndFFor}{\unskip\ \algorithmicend\ \algorithmicfor}

\makeatletter
\newcommand{\subalign}[1]{%
  \vcenter{%
    \Let@ \restore@math@cr \default@tag
    \baselineskip\fontdimen10 \scriptfont\tw@
    \advance\baselineskip\fontdimen12 \scriptfont\tw@
    \lineskip\thr@@\fontdimen8 \scriptfont\thr@@
    \lineskiplimit\lineskip
    \ialign{\hfil$\m@th\scriptstyle##$&$\m@th\scriptstyle{}##$\crcr
      #1\crcr
    }%
  }
}
\makeatother

\newtheorem{frm-def}{Definition}
\newtheorem{frm-prop}{Proposition}
\newtheorem{frm-lemma}{Lemma}
\newtheorem{frm-def-sup}{Definition}
\newtheorem{frm-prop-sup}{Proposition}
\newtheorem{frm-lemma-sup}{Lemma}

\title{Encapsulating models and approximate inference programs in probabilistic modules}

\author{
  Marco F.~Cusumano-Towner\\
  \small Computer Science \& Artificial Intelligence Laboratory\\
  \small Massachusetts Institute of Technology\\
  \texttt{marcoct@mit.edu} \\
  \And
  Vikash K.~Mansinghka\\
  \small Department of Brain \& Cognitive Sciences\\
  \small Massachusetts Institute of Technology\\
  \texttt{vkm@mit.edu} \\
}

\begin{document}

\maketitle
\vspace{-5mm}

\begin{abstract}
This paper introduces the probabilistic module interface, which allows encapsulation of complex probabilistic models with latent variables alongside custom stochastic approximate inference machinery, and provides a platform-agnostic abstraction barrier separating the model internals from the host probabilistic inference system. The interface can be seen as a stochastic generalization of a standard simulation and density interface for probabilistic primitives. We show that sound approximate inference algorithms can be constructed for networks of probabilistic modules, and we demonstrate that the interface can be implemented using learned stochastic inference networks and MCMC and SMC approximate inference programs.
\end{abstract}

\section{Introduction}
We present the probabilistic module interface, which allows encapsulation of complex latent variable models with custom stochastic approximate inference machinery. The modules interface can be seen as a generalization of previously proposed interfaces for ``elementary'' random procedures in probabilistic programming languages: it does not require the module author to specify a marginal input-output density. Instead, module authors are only obligated to (i) provide a way to stochastically ``regenerate'' traces of the internal latent variables, subject to constraints on the module's output, and (ii) provide a way to calculate a weight for this regeneration. We show this is sufficient for constructing sound approximate inference algorithms over networks of modules, including a Metropolis-Hastings procedure that can be seen as the module-level analogue of the single-site Metropolis-Hastings procedures that are commonly used with ``lightweight'' implementations of probabilistic programming languages \cite{goodman2012church}, \cite{wingate2011lightweight}.

This paper illustrates module networks by defining the mathematical interface and providing an example application to linear regression with outliers. This application contains two modules: (i) a complex prior over a binary ``model selection'' variable determining the prior prevalence of outliers, using a learned bottom-up network for regeneration, and (ii) a linear regression model with binary outlier indicators, using sequential Monte Carlo for regeneration of the outlier indicators (thereby avoiding an exponential sum over all possible indicator settings).

\begin{figure}[h]
\begin{subfigure}[b]{0.40\textwidth}
    \includegraphics[width=\textwidth]{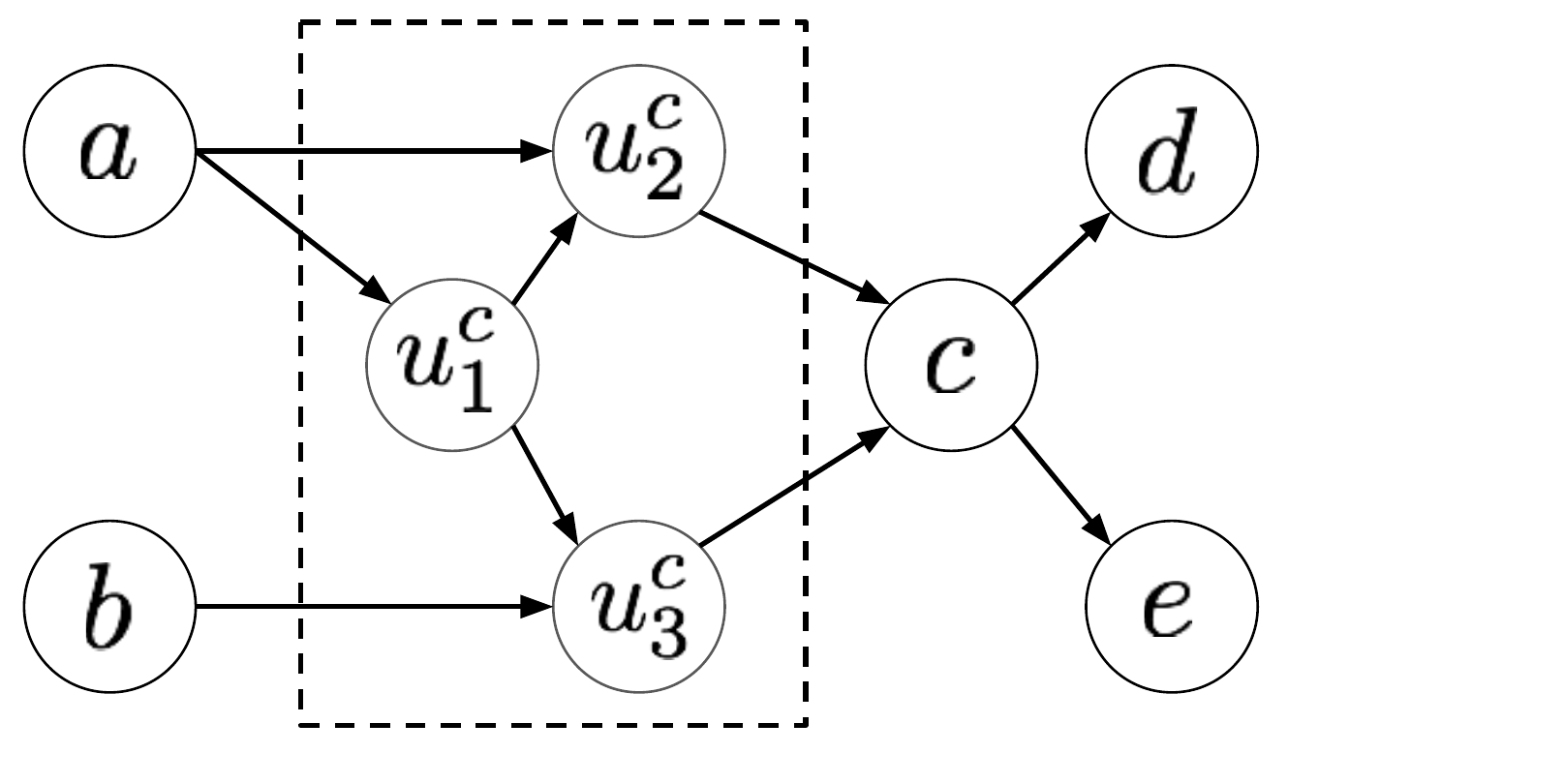}
    \caption{~}
    \label{fig:encapsulate_a}
\end{subfigure}
\begin{subfigure}[b]{0.35\textwidth}
    \includegraphics[width=\textwidth]{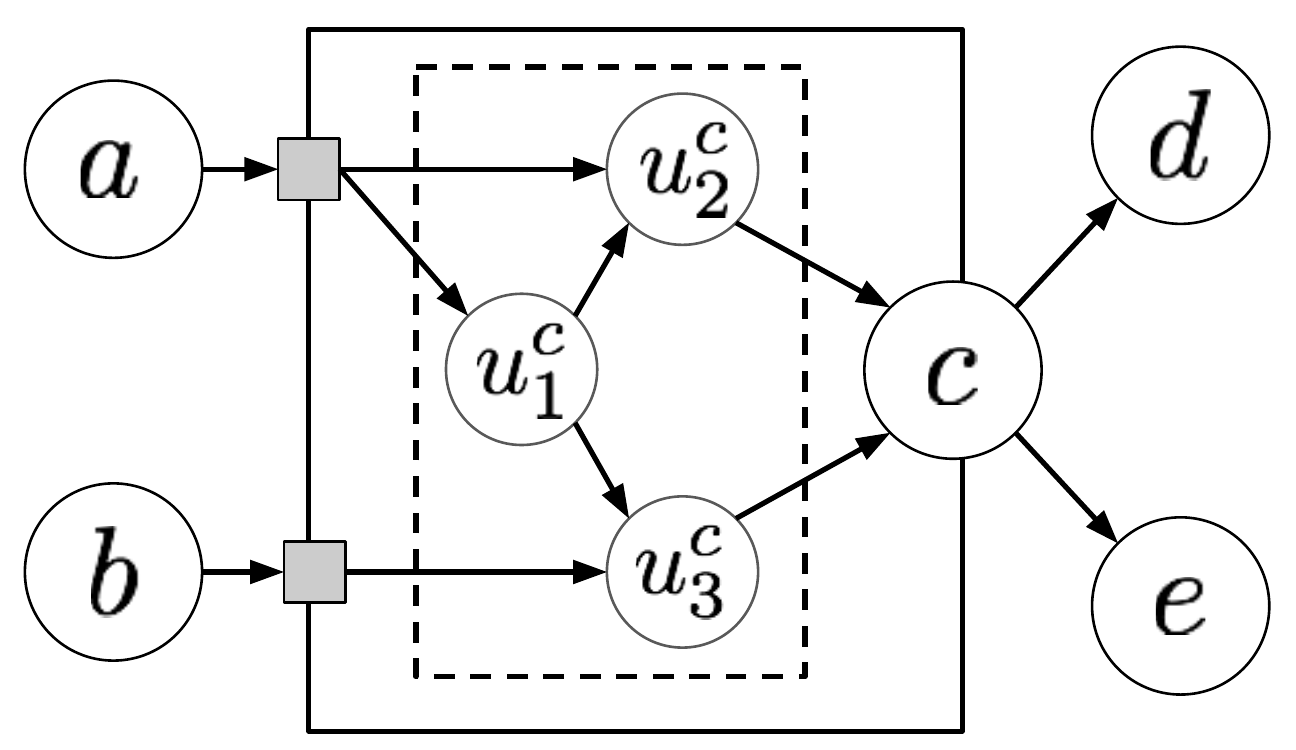}
    \caption{~}
    \label{fig:encapsulate_b}
\end{subfigure}\hfill
\begin{subfigure}[b]{0.18\textwidth}
    \includegraphics[width=\textwidth]{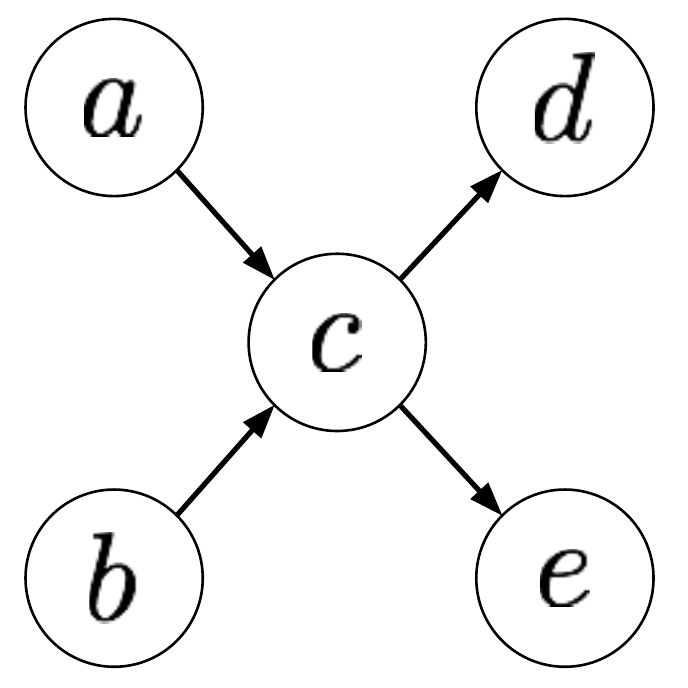}
    \caption{~}
    \label{fig:encapsulate_c}
\end{subfigure}\hfill
\caption{Encapsulating latent variables in a probabilistic model into a probabilistic module: (a) shows the original probabilistic model, with the latent variables $u^c_1,u^c_2,u^c_3$ that are to be abstracted away in a dashed box. (b) shows the model in which the latent variables $u^c_1,u^c_2,u^c_3$ have been made into the internal auxiliary variables $u := (u^c_1,u^c_2,u^c_3)$ of a probabilistic module with output $z := c$ and input $x := (a, b)$. (c) shows the declarative semantics of the resulting module network. }
\label{fig:encapsulate}
\end{figure}

\section{The probabilistic module interface}
In several existing probabilistic programming systems\footnote{Univariate versions of this interface are used in \cite{wood2014new} and \cite{dippl}}  \cite{goodman2012church}, \cite{mansinghka2014venture}, \cite{mansinghka2015bayesdb}, \cite{saad2016probabilistic}, probabilistic modeling primitives implement a simulator procedure ($\mathtt{simulate}$) which samples outputs $z$ given inputs $x$ from a distribution $p(z;x)$ and log-density evaluation procedure ($\mathtt{logpdf}$), which evaluates $\log p(z;x)$. Together, these two procedures enable inference programs to run valid approximate inference algorithms such as MCMC and SMC over the composite probabilistic model. This interface is summarized in Figure~\ref{fig:interfaces}a.

We propose a stochastic generalization of this interface, called the \emph{probabilistic modules interface}, that replaces $\mathtt{logpdf}$ with a stochastic generalization called $\mathtt{regenerate}$. Unlike the $\mathtt{simulate}$ and $\mathtt{logpdf}$ interface, the probabilistic modules interface, summarized in Figure~\ref{fig:interfaces}b, is able to represent probabilistic computations $u,z \sim p(u,z;x)$ that involve internal `auxiliary' random variables $u$ that cannot be exactly marginalized out to compute the log-density $p(z;x) = \sum_u p(u,z;x)$. This is made possible by implementing a sampler that samples values for the auxiliary variables $u$ given inputs $x$ and outputs $z$ from a \emph{regeneration distribution}, which is denoted $q(u;x,z)$.

\begin{figure}[h]
\begin{tabular}{c|c}
$z \gets \mathtt{simulate}(x) \; \mbox{for} \; z \sim p(z;x)$ & $\left(z, \log \displaystyle\frac{p(u,z;x)}{q(u;x,z)}\right) \gets \mathtt{simulate}(x) \; \mbox{for} \; u, z | x \sim p(u, z;x)$\\
[2mm]\\
$\log p(z;x) \gets \mathtt{logpdf}(x,z)$ & $\log \displaystyle\frac{p(u,z;x)}{q(u;x,z)} \gets \mathtt{regenerate}(x,z) \; \mbox{for} \; u | x, z \sim q(u;x,z)$\\
(a) & (b)
\end{tabular}
\caption{(a) shows the standard $\mathtt{simulate}$ and $\mathtt{logpdf}$ interface for elementary random procedures, with inputs $x$ and outputs $z$. (b) shows its stochastic generalization, the probabilistic modules interface, with module inputs $x$, module auxiliary variables $u$ and module outputs $z$.}
\label{fig:interfaces}
\end{figure}

If there are no auxiliary variables $u$, then the $\mathtt{regenerate}$ procedure reduces to the deterministic $\mathtt{logpdf}$ procedure. In the presence of auxiliary variables $u$, $\mathtt{regenerate}$ may be understood as using an unbiased single-sample importance sampling estimate of the output probability, where $q(u;x,z)$ is the importance distribution:
$\mbox{E}_{u | x, z \sim q(u;z,x)} \left[ p(u,z;x) / q(u;z,x) \right] = p(z;x)$.
Indeed, in the extreme setting in which the regeneration distribution is identical to the conditional distribution on auxiliary variables given inputs and outputs ($q(u;x,z) = p(u|x,z)$), this estimate is deterministic and exact, and $\mathtt{regenerate}$ is again identical to $\mathtt{logpdf}$.
Finally, note that the probabilistic modules interface does not require the auxiliary variables $u$ to be stored in memory all at once. This is useful when the log-weight $\log (p(u,z;x) / q(u;x,z))$ can be incrementally computed during sampling of $u$ from $p(u,z;x)$ and $q(u;x,z)$. Such cases are discussed in Section~\ref{sec:encapsulate}.

\section{Implementing MCMC over probabilistic module networks}
When we compose probabilistic modules in a directed acyclic graph, the resulting \emph{probabilistic module network} has the same declarative semantics as a Bayesian network with nodes for module outputs $z_i$. Both the module network and the Bayesian network represent the joint distribution on module outputs, with any module auxiliary variables $u_i$ marginalized out. The existence of auxiliary variables in the modules only changes how approximate inference is performed.

Valid MCMC algorithms can easily be constructed over probabilistic module networks. In fact existing standard Metropolis-Hastings (MH) algorithms for inference in Bayesian networks need only a slight modification for use with modules (see Algorithm~\ref{alg:modules_mh}). The only change required is the storage of the current log-weight for each probabilistic module. The current log-weight for module $i$ is accessed with \textproc{lookup-log-weight}$(i)$ and updated with \textproc{update-log-weight}$(i,\ell)$ during MCMC inference. These values are initialized by running $\mathtt{simulate}$ for each module whose output is not observed and $\mathtt{regenerate}$ for each module whose output is oberved, following a topological ordering of nodes in the network. Note that for single-site MH in a Bayesian network, the \textproc{lookup-log-weight} call and the $\mathtt{regenerate}$ call of Algorithm~\ref{alg:modules_mh} are replaced with $\mathtt{logpdf}$.
A Markov chain constructed from mixtures and cycles of the update of Algorithm~\ref{alg:modules_mh} admits the posterior as a marginal of its stationary distribution, which is defined on the space of all unobserved module outputs $z$, and all module auxiliary variables $u$.

\begin{algorithm}[H]
    \begin{algorithmic}[1]
    \Require Module $i$ whose output to update, proposal distribution $r(z_i';z_i)$, previous value $z_i$.
    \State $z_i' \sim r(\cdot; z_i)$ \Comment{Sample proposed value for module $i$ output}
    \For{$j \in \Call{children}{i} \cup \{i\}$}
        \State $\ell_j \gets \Call{lookup-log-weight}{j}$ \Comment{Look up previous log-weight for module $j$}
        \State $\ell'_j \gets j.\mathtt{regenerate}(x_j',z_j')$ \Comment{Estimate $\log p(z_j';x_j')$ using value $z_i'$}
    \EndFor
    \State $s \gets \mbox{Uniform}(0,1)$
    \If{$\log s \le \log r(z_i;z_i') - \log r(z_i';z_i) + \sum_{j \in \Call{children}{i} \cup \{i\}} (\ell'_j - \ell_j)$}
        \State $z_i \gets z_i'$ \Comment{Accept the proposal}
        \For{$j \in \Call{children}{i} \cup \{i\}$}
            \State $\Call{update-log-weight}{j, \ell'_j}$
        \EndFor
    \EndIf
    \end{algorithmic}
    \caption{Single-site Metropolis-Hastings (MH) update in probabilistic module network} \label{alg:modules_mh}
\end{algorithm}

\section{Encapsulating models and inference programs in probabilistic modules} \label{sec:encapsulate}
We now show how to encapsulate a probabilistic model $p(v,z;x)$ with internal latents $v$ and outputs $z$ as a probabilistic module with the declarative semantics of the marginal distribution on outputs $p(z;x)$, as shown in Figure~\ref{fig:encapsulate}. This is useful if $v$ is high dimensional or analytically intractable, and we are unable to implement $\mathtt{logpdf}$ by marginalizing out $v$ exactly.

\begin{figure}[h]
\centering
\begin{minipage}{1.0\textwidth}
\begin{minipage}{.5\textwidth}
\renewcommand{\thesubfigure}{a}
\begin{subfigure}[b]{1.0\textwidth}
    \centering
    \includegraphics[width=\textwidth]{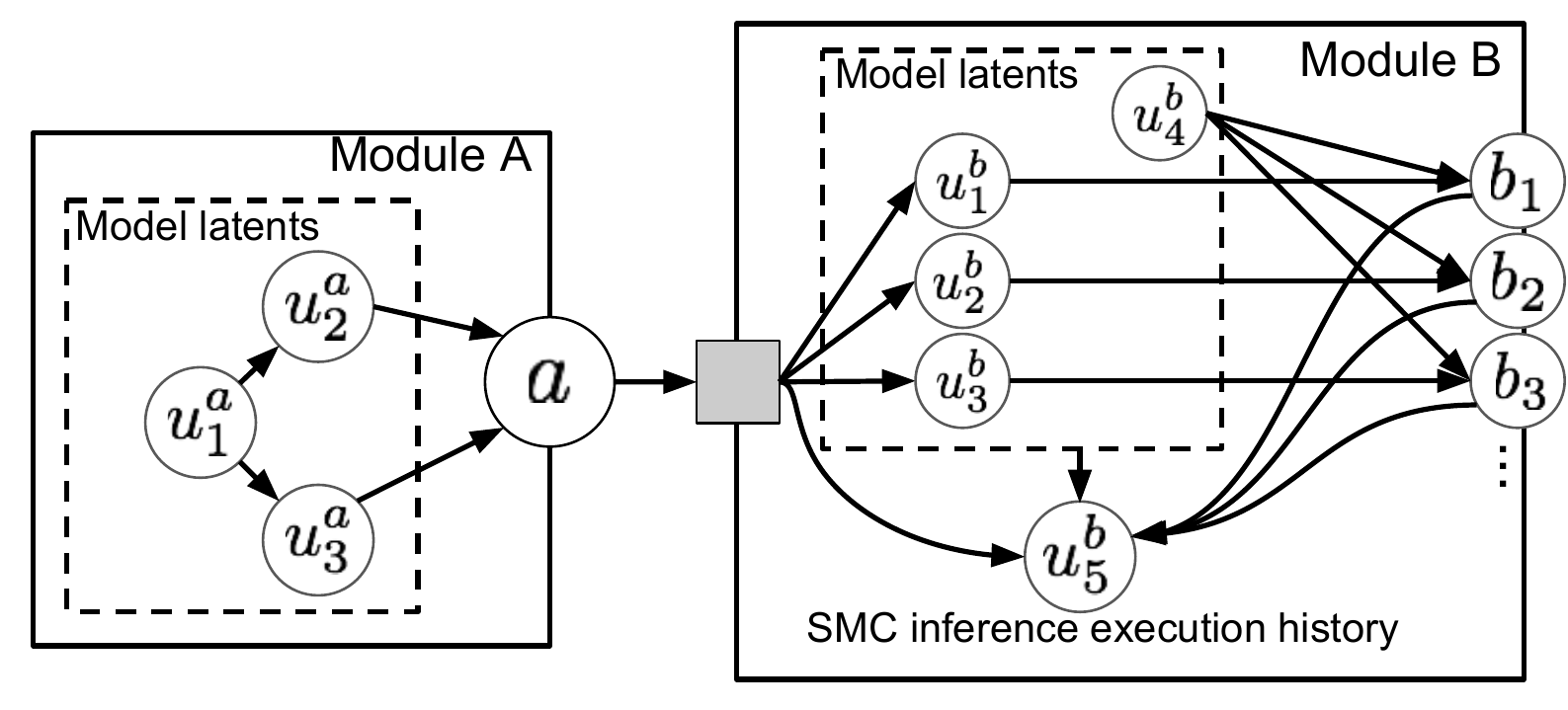}
    \caption{Data flow of forward module network simulation.}
    \label{fig:example_modules}
\end{subfigure}\\
\end{minipage}%
\begin{minipage}{.5\textwidth}
\renewcommand{\thesubfigure}{c}
\begin{subfigure}[t]{0.7\textwidth}
    \centering
    \includegraphics[width=\textwidth]{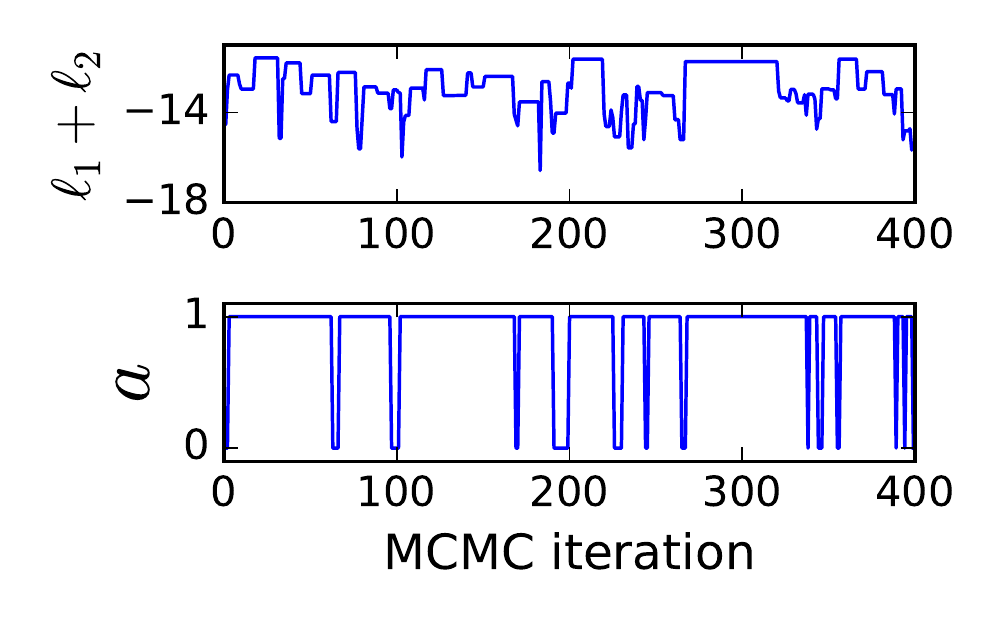}
    \caption{~}
    \label{fig:log_weight_trace}
\end{subfigure}%
\renewcommand{\thesubfigure}{d}
\begin{subfigure}[t]{0.29\textwidth}
    \centering
    \includegraphics[width=\textwidth]{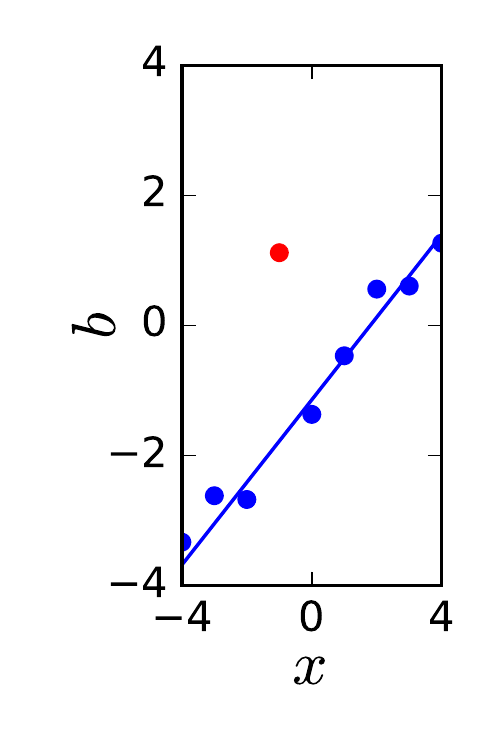}
    \caption{~}
    \label{fig:regression_rendering}
\end{subfigure}
\end{minipage}\\
\begin{minipage}{0.5\textwidth}
\renewcommand{\thesubfigure}{b}
\begin{subfigure}[b]{1.0\textwidth}
    \centering
    \includegraphics[width=\textwidth]{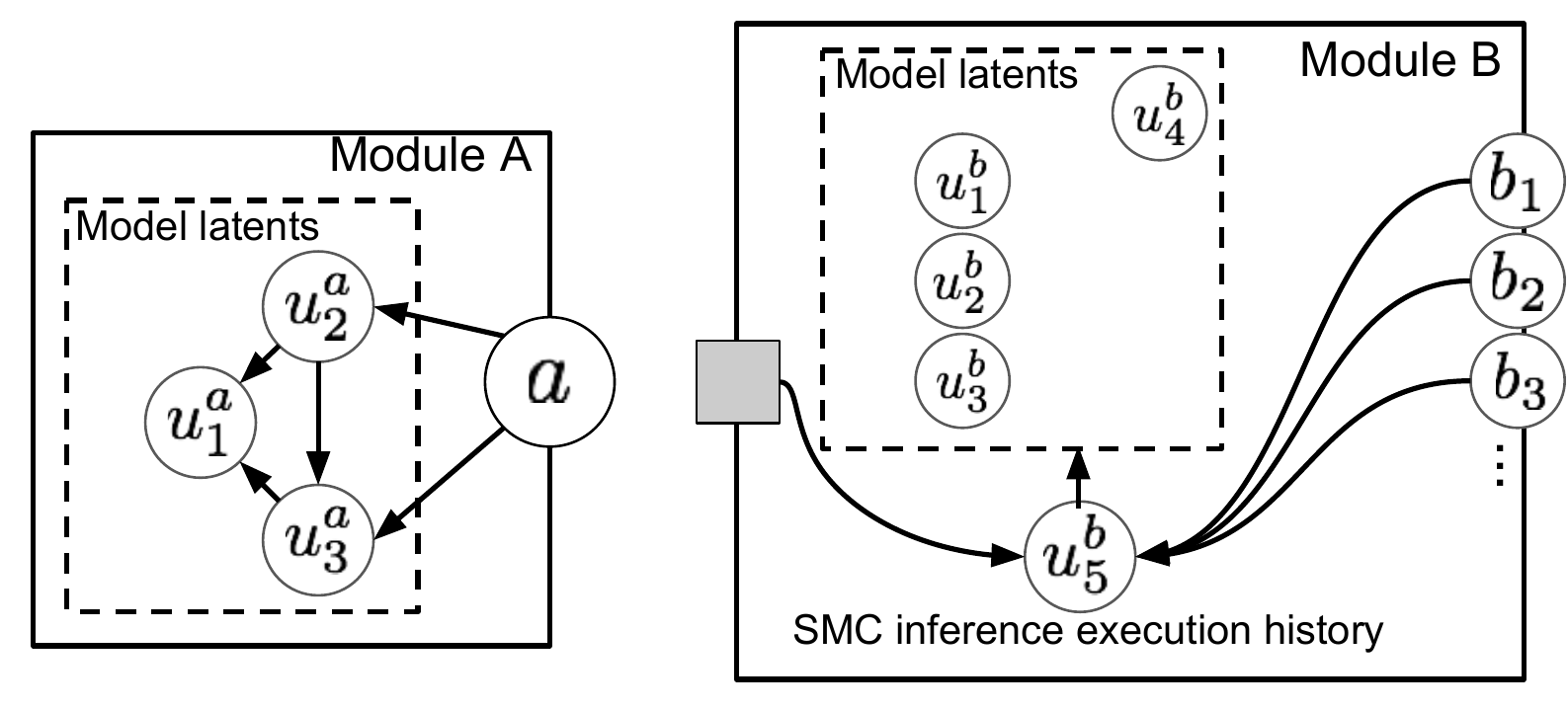}
    \caption{Data flows of regeneration for the two modules.}
    \label{fig:regen_dataflows}
\end{subfigure}
\end{minipage}%
\begin{minipage}{.5\textwidth}
\renewcommand{\thesubfigure}{e}
\begin{subfigure}[b]{1.0\textwidth}
    \begingroup
    \footnotesize\begin{align*}
        u^b_4 &\sim \mathcal{N}\left(\bm{\mu}{=}\left[\begin{array}{c}0\\0\end{array}\right], \Sigma{=}\left[\begin{array}{cc}1 & 0\\0 & 0.25 \end{array}\right]\right)\\
        u^b_i &\sim \left\{ \begin{array}{ll} \mbox{Bernoulli}(0.01) \mbox{ if } a = 0\\\mbox{Bernoulli}(0.1) \mbox{ if } a = 1  \end{array}\right.\\
        b_i &\sim \left\{ \begin{array}{ll} \mathcal{N}(\mu{=}\bm{\mu}^T [1, x_i], \sigma{=}0.22) \mbox{ if } u^b_i{=}0\\ \mathcal{N}(\mu{=}\bm{\mu}^T [1, x_i], \sigma{=}3.16) \mbox{ if } u^b_i = 1 \end{array} \right.\\
        & \mbox{ for } i=1,2,\ldots,9
    \end{align*}
    \endgroup
    \caption{Internal probabilistic model of module $B$}
    \label{fig:math}
\end{subfigure}
\end{minipage}
\end{minipage}
\caption{Illustration of module networks on an example application to linear regression with outliers. (a) shows two latent variable models encapsulated in probabilistic modules and composed in a probabilistic module network. Module $A$ encodes a prior distribution on $a \in \{0,1\}$, which determines the prior prevalence of outliers. Module $B$ encodes the linear regression outlier model.  To treat modules like a single node in a Bayesian network when we lack an their marginal output density, we perform stochastic inversion (or ``regeneration'') of the module. 
(b) shows the data flow of regeneration for the two modules.  Module $A$ uses a learned stochastic inverse network and module $B$ uses sequential Monte Carlo (SMC) for regeneration. Despite being approximate, these permit valid Metropolis-Hastings over the exposed latent variable $a$, as if we had the marginal output densities of the two modules.
 (c) shows traces of $a$ and the total log-weight for an MCMC run using Algorithm~\ref{alg:modules_mh} in this model, with observed values for $b_1,b_2,\ldots,b_9$. The total log-weight varies stochastically even when $a$ is static.
 (d) shows a rendering of the latent variables of the model encapsulated in module $B$ (the line and outlier statuses) and the dataset at a point in the chain
 (e) shows the internal probabilistic model of module $B$.
 }
\label{fig:example}
\end{figure}

We begin by defining the module auxiliary variables as the model's internal latents ($u := v$). Then, the probabilistic module interface requires us to construct a sampler for the regeneration distribution $v|x, z \sim q(v;x,z)$ where $q(v;x,z)$ is an approximation to $p(v|z;x)$ such that we can efficiently compute the log-weight $\log (p(v,z;x) / q(v;x,z))$. It is sometimes possible to \emph{learn} a sampler $q(v;x,z)$ (see \cite{morris2001recognition} for a pioneering example of this approach, and the `stochastic inverses' of \cite{stuhlmuller2013learning}) or to learn the model $p(v,z;x)$ and the regeneration sampler at the same time (e.g. the generative models and recognition networks of \cite{kingma2013auto}) such that the log-weight is tractable. We illustrate this approach for Module $A$ of Figure~\ref{fig:example_modules}, which uses a learned stochastic inverse network trained using samples from the prior of the model as described in \cite{stuhlmuller2013learning}. The log-weight is tractable because the learned $q(v;z,x)$ contains no additional random variables beyond those in the model itself ($v$).

However, if we wish to use generally applicable stochastic inference programs implementing MCMC \cite{andrieu2003introduction} and sequential Monte Carlo (SMC) \cite{del2006sequential} for the regeneration distribution $q(v;x,z)$, it is not possible to compute $\log (p(v,z;x) / q(v;x,z))$ because the marginal output density of the stochastic inference program is intractable.
To handle these cases, we augment the auxiliary variables of the module to include the \emph{execution history} $w$ of the stochastic inference program ($u := (v,w)$). We define the distribution sampled by $\mathtt{regenerate}$ as the joint distribution of the stochastic inference program over its execution history $w$ and output $v$, denoted $q(u;x,z) := q(w,v;x,z)$. We then extend the distribution sampled by $\mathtt{simulate}$ to also sample an execution history $w$ alonside the model latents $v$, using a `meta-inference' program \cite{cusumano2016quantifying} that samples inference execution history given inference output $v$ from a distribution $m(w;x,v,z)$ that approximates the conditional distribution on inference execution histories $q(w|v;x,z)$, so that $p(u,z;x) := p(v,z;x) m(v;x,v,z)$.

As shown in \cite{cusumano2016quantifying} and \cite{cusumanotowner2016smc}, it is possible to construct meta-inference programs for sequential variants of MCMC using detailed balance transition kernels and for multiple-particle SMC with optional detailed balance transition kernels such that the log-weight $\log (p(v,z;x) m(w;x,v,z) / q(w,v;z,x))$ can be efficiently computed on the fly when sampling from $q(w,v;z,x)$ and $p(v,z;x) m(w;x,v,z)$. As the accuracy of $m(w;x,v,z)$ improves (as happens when the number of particles in SMC increases) the log-weights sampled from the probabilistic module converge to the log density $p(z;x)$. Module $B$ of Figure~\ref{fig:example} uses SMC for $q(w,v;z,x)$.

\subsubsection*{Acknowledgements}
This research was supported by DARPA (PPAML program, contract number
FA8750-14-2-0004), IARPA (under research contract 2015-15061000003), the Office
of Naval Research (under  research  contract N000141310333), the Army Research
Office (under agreement number W911NF-13-1-0212),  and gifts from Analog
Devices and Google.
This research was conducted with Government support under and awarded by DoD, Air Force Office of Scientific Research, National Defense Science and Engineering Graduate (NDSEG) Fellowship, 32 CFR 168a.

\bibliographystyle{unsrt}
\bibliography{modules_references} 

\newpage
\appendix

\section{Appendix A: Deriving single-site Metropolis Hastings in module network}
Consider a network of probabilistic modules indexed by $V = \{1, \ldots, N\}$. Let $u_i$ denote the internal auxiliary variables or module $i$, let $z_i$ denote the output of module $i$, and let $x_i$ denote the inputs to module $i$. Each module input $x_i$ is a tuple of module outpus $z_j$ for $j \in \pi_i$, where $\pi_i$ is the sequence of parent module indices of module $i$. Let $c_i$ denote the set of children of module $i$: $(c_i = \{j \in V : i \in \pi_j\})$. Let $p_i(u_i, z_i; x_i)$ denote the simulation distribution for module $i$ and let $q(u_i; x_i, z_i)$ denote the regeneration distribution. Define the collection of all module auxiliary variables by $u = (u_1, \ldots, u_N)$ and the collection of all module outputs by $z = (z_1, \ldots, z_N)$. Define the joint simulation distribution over the network as $p(u, z) := \prod_{i \in V} p_i(u_i, z_i; x_i)$. The declarative semantics of the module network are derived from the marginal distribution over outputs $p(z) := \sum_u p(u, z)$. Suppose a subset of the modules $O \subsetneq V$ are observed, meaning their output is constrained to a value $z_i$. The target distribution of the network is then defined by $p((z_j)_{j \not \in O} | (z_i)_{i \in O})$. We will derive a Metropolis-Hastings algorithm for which the stationary distribution is the distribution $p(u, (z_j)_{j \not \in O} | (z_i)_{i \in O})$. If the algorithm converges to this joint distribution, then the marginal over only $(z_j)_{j \not \in O}$ converges to the network's target distribution.

Consider a standard single-site Metropolis-Hastings update targeting the distribution $p(u, (z_j)_{j \not \in O} | (z_i)_{i \in O})$ where we propose a new value for the output of some module $i \not \in O$. Let $u := (u_1, \ldots, u_N)$ and $z := (z_1, \ldots, z_N)$ denote the state prior to the update. We propose a new value $z_i' \sim r(\cdot; z_i)$, and then propose $u_i' | z_i' \sim q_i(\cdot; x_i, z_i')$. We also propose $u_j | z_i' \sim q_j(\cdot; x_j', z_j)$ for all modules $j \in c_i$, where $x_j'$ includes the updated value $z_i'$. We perform an MH accept/reject step using this proposal, and with the `local' posterior $p(u_i, z_i, (u_j)_{j \in c_i} | x_i, (x_j)_{j \in c_i}, (z_j)_{j \in c_i})$ as the target distribution. The acceptance ratio is:
\begin{align}
\alpha &= \frac{p_i(u_i', z_i'; x_i) \prod_{j \in c_i} p_j(u_j', z_j; x_j')}{p_i(u_i, z_i; x_i) \prod_{j \in c_i} p_j(u_j, z_j; x_j)} \cdot \frac{r(z_i) q_i(u_i; x_i, z_i) \prod_{j \in c_i} q_j(u_j; x_j, z_j)}{r(z_i') q_i(u_i'; x_i, z_i') \prod_{j \in c_i} q(u_j'; x_j', z_j)}
\end{align}
The log-acceptance ratio is:
\begin{align}
\log \alpha = &\log \frac{p_i(u_i', z_i'; x_i)}{q_i(u_i'; x_i, z_i')} - \log \frac{p_i(u_i, z_i; x_i)}{q_i(u_i; x_i, z_i)}\\&+ \sum_{j \in c_i} \left( \log \frac{p_j(u_j',z_j;x_j')}{q_j(u_j',z_j;x_j')} - \log \frac{p_j(u_j,z_j;x_j)}{q_j(u_j,z_j;x_j)} \right)\\ &+ \log r(z_i) - \log r(z_i')
\end{align}
This is the acceptance ratio used in Algorithm~\ref{alg:modules_mh}. Therefore Algorithm~\ref{alg:modules_mh} corresponds to a valid MH update that can be used to compose valid MH algorithms that converge to the posterior $p(u, (z_j)_{j \not \in O} | (z_i)_{i \in O})$, such as a single-site random-scan mixture over Algorithm~\ref{alg:modules_mh} applications for all unobserved modules $i \not \in O$.

\end{document}